\documentclass[10pt,twocolumn,letterpaper]{article}

\usepackage{cvpr}
\usepackage{times}
\usepackage{epsfig}
\usepackage{graphicx}
\usepackage{amsmath}
\usepackage{amssymb}

\usepackage{color,colortbl}
\definecolor{Gray}{gray}{0.90}
\newcolumntype{a}{>{\columncolor{Gray}}c}
\usepackage[utf8]{inputenc} % allow utf-8 input
\usepackage[T1]{fontenc}    % use 8-bit T1 fonts
\usepackage{url}
\usepackage{booktabs}       % professional-quality tables
\usepackage{amsfonts}       % blackboard math symbols
\usepackage{nicefrac}       % compact symbols for 1/2, etc.
\usepackage{microtype}      % microtypography
\usepackage{bbm}
\usepackage{wrapfig}
\usepackage{dblfloatfix}
\usepackage{balance}

\usepackage[pagebackref=true,breaklinks=true,letterpaper=true,colorlinks,bookmarks=false]{hyperref}

\cvprfinalcopy % *** Uncomment this line for the final submission

\def\cvprPaperID{2228} % *** Enter the CVPR Paper ID here

\ifcvprfinal\fi
\begin{document}
	
	%%%%%%%%% TITLE
	\title{Domain Adaptation with Auxiliary Target Domain-Oriented Classifier}
	
	\author{Jian Liang $^1$\qquad 
		Dapeng Hu $^1$\qquad
		Jiashi Feng $^{2,1}$ \\
		$^1$ National University of Singapore (NUS)
		\qquad
		$^2$ Sea AI Lab (SAIL) \\
		{\tt\small liangjian92@gmail.com}\qquad
		{\tt\small dapeng.hu@u.nus.edu}\qquad
		{\tt\small elefjia@nus.edu.sg}
	}
	
	\maketitle
	\thispagestyle{empty}
	
	%%%%%%%%% ABSTRACT
	\begin{abstract}
		Domain adaptation (DA) aims to transfer knowledge from a label-rich but heterogeneous domain to a label-scare domain, which alleviates the labeling efforts and attracts considerable attention.
		Different from previous methods focusing on learning domain-invariant feature representations, some recent methods present generic semi-supervised learning (SSL) techniques and directly apply them to DA tasks, even achieving competitive performance.
		One of the most popular SSL techniques is pseudo-labeling that assigns pseudo labels for each unlabeled data via the classifier trained by labeled data.
		However, it ignores the distribution shift in DA problems and is inevitably biased to source data.
		To address this issue, we propose a new pseudo-labeling framework called Auxiliary Target Domain-Oriented Classifier (ATDOC).
		ATDOC alleviates the classifier bias by introducing an auxiliary classifier for target data only, to improve the quality of pseudo labels. 
		Specifically, we employ the memory mechanism and develop two types of non-parametric classifiers, i.e. the nearest centroid classifier and neighborhood aggregation, without introducing any additional network parameters.
		Despite its simplicity in a pseudo classification objective, ATDOC with neighborhood aggregation significantly outperforms domain alignment techniques and prior SSL techniques on a large variety of DA benchmarks and even scare-labeled SSL tasks.   
	\end{abstract}
	
	\section{Introduction}
	Despite remarkable progress in classification tasks over the past decades, deep neural network models still suffer poor generalization performance to another new domain e.g. classifying real-world object images using a classification model trained on simulated object images \cite{peng2017visda}, due to the well-known dataset shift \cite{quionero2009dataset} or domain shift \cite{tommasi2016learning} problem.
	Hence, to avoid expensive human labeling and utilize prior related labeled datasets, lots of research efforts have been devoted to developing domain adaptation (DA) methods \cite{gong2012geodesic,ganin2016domain,hoffman2016fcns,tsai2018learning,saito2019semi,jiang2020bidirectional,li2020online} to transfer knowledge in the label-rich dataset to a label-scare scenario. 
	Depending on the availability of labeled data in the target domain, one can further divide existing DA methods into two categories, i.e., unsupervised DA \cite{ganin2016domain} and semi-supervised DA \cite{saito2019semi}. 

	This paper mainly focuses on unsupervised DA where no labeled data is available in the target domain during training.
	Recently, deep unsupervised DA approaches have almost dominated this field with promising results~\cite{long2015learning,ganin2016domain,long2018conditional,lee2019sliced,kang2019contrastive,cicek2019unsupervised}, and most of them focus on learning domain-invariant feature representations that achieve a small error on the source domain at the same time.
	With the assumption about covariate shift in ~\cite{ben2010theory}, the learned representations together with the classifier built on the source domain are able to generalize well to the target domain.
	However, the strict assumption does not always hold in real-world applications. 
	Another line of research ignores transferable representation learning but directly resorts to semi-supervised learning (SSL) techniques for the DA problems \cite{french2018self,chen2019domain,rukhovich2019mixmatch,cui2020towards,jin2020less}, where the target domain could be readily treated as the unlabeled set in SSL.
	For instance, MixMatch \cite{berthelot2019mixmatch}, a popular SSL approach, has been successfully applied by \cite{rukhovich2019mixmatch} that wins prizes of the \emph{VisDA 2019 Challenge}.
	But these SSL-based DA methods may fail to classify target samples far away from the source domain due to the ignorance of domain shift.
	
	One of the most popular SSL techniques\textemdash pseudo labeling~\cite{lee2013pseudo} iteratively assigns pseudo labels corresponding to the maximum prediction scores for each unlabeled data in the target domain and then retrains the network with the pseudo-labeled data in a supervised manner.
	However, the network is inevitably biased to the labeled source data during training, giving low-quality pseudo labels and propagating errors in the target domain.
	To tackle this issue, we propose a new pseudo-labeling framework termed Auxiliary Target Domain-Oriented Classifier (ATDOC) for DA problems.
	Generally, ATDOC attempts to alleviate the labeling bias by introducing an auxiliary classifier for target data only.
	In particular, we design two types of non-parametric classifiers, i.e., nearest centroid classifier (NC) and neighborhood aggregation (NA), to avoid additional network parameters.
	Both class centroids and local neighborhood structures are capable of representing the target domain, thus the generated target-oriented pseudo labels are fairly unbiased and reliable.
	
	\begin{figure*}[t]
		\centering
		\includegraphics[width=0.85\textwidth]{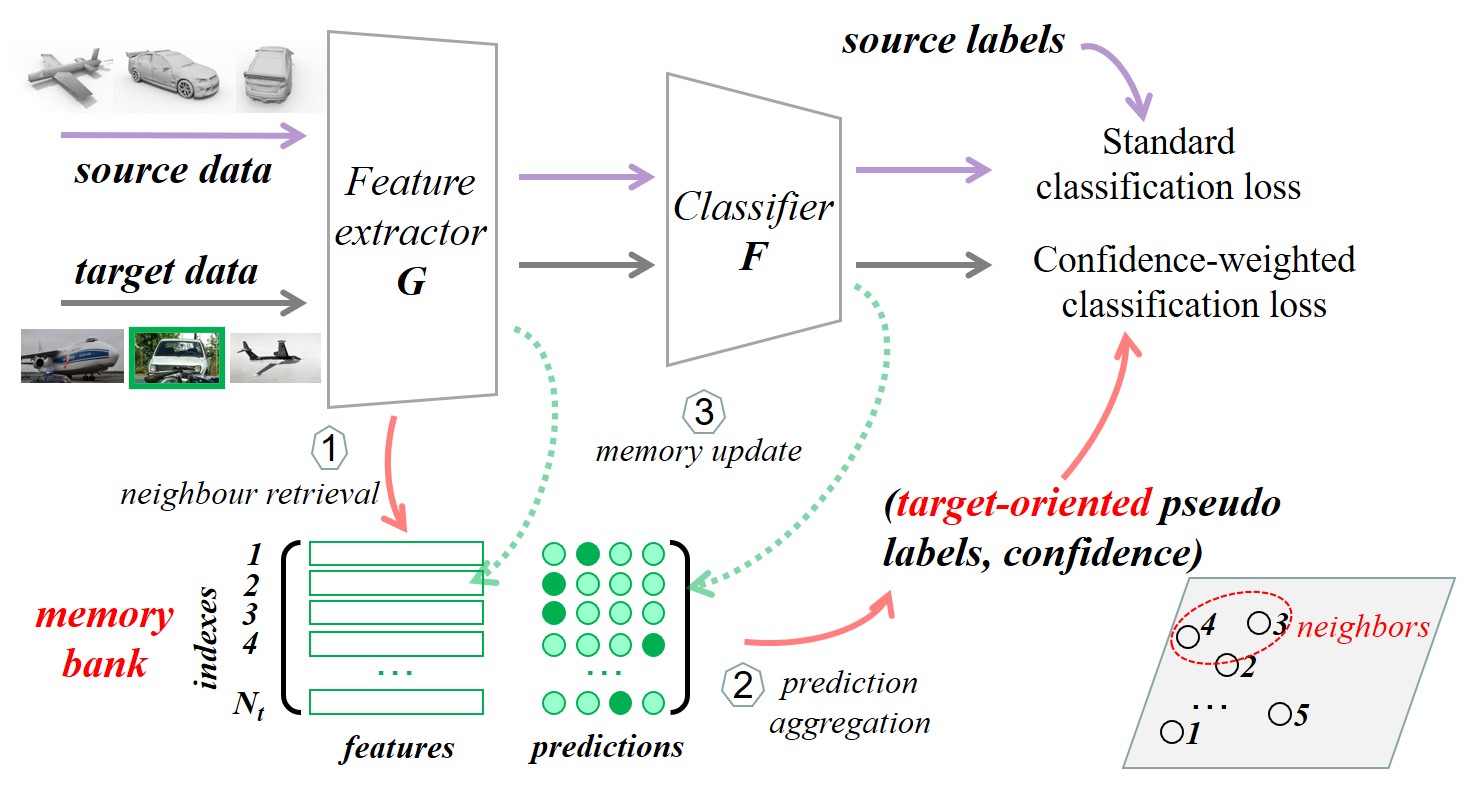}
		\vspace{3pt}
		\caption{The pipeline of our proposed framework ATDOC with neighborhood aggregation (NA) for UDA. Different from existing methods that mostly rely on feature-level domain alignment, ATDOC addresses domain shift by alleviating the classifier bias via an auxiliary classifier for target data only during adaptation. ATDOC-NA employs a memory bank and develops neighborhood aggregation to help build the domain-specific classifier $\mathbf{F}_t$, and expects to generate unbiased accurate pseudo labels along with confidence weights for unlabeled data.}
		\label{fig:framework}
	\end{figure*}
	
	To enable global structure learning with mini-batch optimization, we introduce a memory module to store information over all the unlabeled target samples.
	Besides, since no labeled data is available in the target domain, noisy pseudo labels are directly exploited as an alternative.
	Specifically, concerning the NC classifier training, we follow \cite{wen2016discriminative,xie2018learning} and construct a memory bank to store feature representations of the class centroids.
	The exponential moving average strategy is adopted to dynamically update centroids in the memory bank.
	Built on the centroids, the NC classifier readily offers a target-oriented pseudo-label for each unlabeled data. 
	Compared with coarse class centroids, training a NA classifier needs to construct a large memory bank that consists of the features along with soft predictions over all the target samples.
	Through simple aggregation, NA feasibly generates confidence as an instance weight besides the one-hot pseudo label.
	An overview of ATDOC-NA is shown in Fig.~\ref{fig:framework}.
	Generally, the network at first several iterations is still biased to source data, but as the number of training iterations increases, retraining with target-oriented pseudo labels gradually adjusts the network to unlabeled target data, achieving domain alignment via optimizing two classification objectives. 
	
	\textbf{Our contributions} are summarized as follows:
	(i) we propose ATDOC, a new framework to combat classifier bias that provides a new perspective of addressing domain shift. 
	(ii) we exploit the memory bank and develop two types of non-parametric classifiers, not involving complicated network architectures with extra parameters. 
	(iii) despite its simplicity, ATDOC achieves competitive or better results than prior state-of-the-arts under a variety of DA settings, e.g., partial-set unsupervised DA \cite{cao2018partial} and semi-supervised DA.
	ATDOC can be seamlessly integrated into existing domain-invariant feature learning methods and further boost their adaptation performance.
	Besides, we study an SSL setting with only a few annotated data points available and find ATDOC also performs better than other SSL techniques even without domain shift.

	\section{Related Work}
	\subsection{Deep Domain Adaptation} 
	Deep domain adaptation methods \cite{csurka2017comprehensive,kouw2019review,wilson2020survey}. aim to learn more transferable representations by embedding domain adaptation in the pipeline of deep learning. 
	Generally, the weights of the deep architecture containing a feature encoder and a classifier layer are shared for both domains, and various distribution discrepancy measures \cite{tzeng2014deep,long2015learning,ganin2015unsupervised} are developed to promote domain confusion in the feature space.
	Among them, maximum mean discrepancy (MMD) \cite{gretton2007kernel} and $\mathcal{H}\Delta\mathcal{H}$-distance \cite{ben2010theory} are two most favored measures.
	To circumvent the problem that marginal distribution alignment cannot guarantee different domains are semantically aligned, following works \cite{long2018conditional,cicek2019unsupervised} exploit pseudo labels on the target domain to perform conditional distribution alignment.
	The learned classifier still fails to generalize well on the target domain, as it is mainly built on the labeled source data.
	
	Another line of research \cite{long2016unsupervised,rozantsev2018beyond,liang2020we} exploits the individual characteristics of each domain by dropping the weight-sharing assumption fully or partially.
	Shu \etal \cite{shu2018dirt} propose non-conservative domain
	adaptation and incrementally refine the preciously learned classification boundary to fit the target domain only.
	With the classifier shared, Tzeng \etal \cite{tzeng2017adversarial} first learn the source feature encoder and then the target feature encoder sequentially. While Bousmalis \etal \cite{bousmalis2016domain} jointly learn the domain-shared encoder and domain-specific private encoders.
	Besides, Chang \etal \cite{chang2019domain} share all other model parameters but specialize batch normalization layers within the feature encoder.
	Liang \etal \cite{liang2020we} learn the target-specific feature extractor while only operating on the hypothesis induced from the source data.
	Long \etal \cite{long2016unsupervised} pursue classifier
	adaptation through a new residual transfer module that bridges the source classifier and target classifier.
	Compared with these methods, ATDOC does not introduce any new layers and aims to learn one shared classifier for both domains with a virtual target-oriented classifier.
	
	\subsection{Semi-supervised Learning with Regularization}
	Semi-supervised learning (SSL) is a classical machine learning approach, which aims to exploit unlabeled data to produce a considerable improvement in learning accuracy.
	Typically, besides the classification objective for labeled data, SSL methods \cite{zhu2005semi,van2020survey} resort to the cluster assumption or low-density separation assumption to fully exploit unlabeled data as regularization, e.g., entropy minimization \cite{grandvalet2005semi}.
	The virtual adversarial training loss~\cite{miyato2018virtual} is proposed to measure local smoothness of the conditional label distribution around each input data point against local perturbation.
	By contrast, temporal ensembling \cite{laine2016temporal} is developed to pursue the consistency regularization over multiple training epochs.
	Further, graph-based techniques e.g. label propagation \cite{zhu2002learning} and manifold
	regularization \cite{belkin2006manifold} are favored by a majority of researchers before the deep learning area, which has been effectively extended in recent works \cite{kipf2016semi,iscen2019label}.
	In addition, data augmentation techniques such as MixUp and random crop are also utilized in MixMatch~\cite{berthelot2019mixmatch} which achieves state-of-the-art results in multiple SSL benchmarks. 
	
	In fact, unsupervised DA can be considered as a special case of transductive SSL where labeled data and unlabeled data are sampled from different distributions.
	Recent studies \cite{chen2019domain,cui2020towards,jin2020less} show that regularization terms on unlabeled data without explicit feature-level domain alignment achieve promising adaptation results.
	In particular, the MaxSquare loss is developed in~\cite{chen2019domain} to prevent the training process from being dominated by easy-to-transfer samples in the target domain.
	In contrast, the diversity of conditional predictions is considered through batch nuclear-norm maximization~\cite{cui2020towards} and class confusion minimization~\cite{jin2020less}, respectively.
	
	\subsection{Pseudo-labeling in Transductive Learning}
	Pseudo-labeling is a heuristic approach to transductive SSL, providing an alternative solution to regularization.
	As a typical example, `Pseudo-Label' \cite{lee2013pseudo} progressively treats high-confidence predictions on unlabeled data as true labels (called pseudo labels) and employ a standard cross-entropy loss during re-training.
	Following works \cite{shi2018transductive,deng2019cluster} incorporate pseudo labels to perform discriminative clustering for features of unlabeled data.
	To alleviate noises in the instance-wise pseudo labels, \cite{iscen2019label} relies on graph-based label propagation and \cite{zou2018unsupervised} considers the class imbalance problem, obtaining better pseudo labels.
	Besides, the weights of different pseudo labels are considered, \cite{choi2019pseudo} adopts a curriculum learning strategy and \cite{zou2019confidence} treats pseudo-labels as continuous latent variables jointly optimized via alternating optimization.
	Our method is different from all such prior work in that pseudo-labels are inferred by an auxiliary parameter-free target-oriented classifier effectively and efficiently.
	
	\subsection{Transductive Learning with Memory Bank}
	A memory bank can be read and written to remember past facts, enabling global structure learning with mini-batch optimization \cite{sukhbaatar2015end}.
	A recent study \cite{chen2018semi} first exploits the memory mechanism in the network training for SSL and computes the memory prediction for each training sample by the key addressing and value reading.
	Inspired by instance discrimination \cite{wu2018unsupervised}, Saito \etal~\cite{saito2020universal} employ a memory
	bank and propose an entropy minimization loss to encourage neighborhood clustering in the target domain.
	Xie \etal \cite{xie2018learning} utilize the memory bank to compute centroids from both domains for centroid matching.
	Besides, Zhong \etal~\cite{zhong2019invariance} leverage an exemplar memory module that saves up-to-date features for target data and computes the invariance learning loss for unlabeled target data.
	Among them, \cite{chen2018semi} is the most closely related work to ours, but \cite{chen2018semi} is merely proposed for SSL that only utilizes the labeled data for memory update and ignores self-learning in the unlabeled data.
	Besides, in contrast to supervised center loss \cite{wen2016discriminative}, the memory bank in the ATDOC framework e.g. storing class centroids is updated in an unsupervised way. 
	
	\section{Methodology}
	For the unsupervised DA (UDA) task, we are given a labeled source domain $\mathcal{D}_s = \{(x_i^s, y_i^s)\}_{i=1}^{N_s}$ with $K$ categories and an unlabeled target domain $\mathcal{D}_{tu} = \{(x_i^t)\}_{i=1}^{N_{tu}}$, 
	while in semi-supervised DA (SSDA), we are given an additional labeled subset of the target domain $\mathcal{D}_{tl} = \{(x_i^t, y_i^t)\}_{i=1}^{N_{tl}}$.
	$\mathcal{D}_{t} = \mathcal{D}_{tu} \cup \mathcal{D}_{tl}$ denotes the entire target domain, and $\mathcal{D}_{tl}$ is empty in UDA.
	This paper mainly focuses on the vanilla closed-set setting where two domains share the same categories.
	The ultimate goal of both UDA and SSDA is to label the target samples in $\mathcal{D}_{tu}$ via training the model on $\mathcal{D}_{s} \cup \mathcal{D}_{t}$.
	
	As shown in Fig.~\ref{fig:framework}, we employ the widely-used architecture \cite{ganin2015unsupervised} which consists of two basic modules, a feature extractor $\mathbf{G}$ and a classifier $\mathbf{F}$. 
	Based on where to align, DA approaches can be roughly categorized into three main cases, i.e., pixel-level~\cite{hoffman2018cycada,sankaranarayanan2018generate}, feature-level~\cite{ganin2015unsupervised,tzeng2017adversarial,long2018conditional,li2020maximum} and output-level~\cite{chen2019domain,cui2020towards,jin2020less}.
	Compared with time-consuming pixel-level transfer methods, feature-level domain alignment is cheap and flexible, attracting the most attention.
	Belonging to the output-level domain alignment, our ATDOC framework well combats the classifier bias and can be seamlessly combined with feature-level DA approaches.
	
	\subsection{A Closer Look at Pseudo-labeling}
	To fully utilize the unlabeled data, following classic self-training \cite{zhu2005semi}, Lee \cite{lee2013pseudo} presents a simple method with training deep neural networks for SSL. It picks up the class $\hat{y}$ with the maximum predicted probability as true labels each time the weights are updated.
	\begin{equation}
		\begin{split}
			\hat{y}_i &= \arg\max\limits_{k} p_{i,k}, \ i=1, 2, \cdots, N_{tu}, \\
			\mathcal{L}_{pl} &= - \frac{\alpha}{N_{tu}} \sum\nolimits_{i=1}^{N_{tu}} \log p_{i,\hat{y}_i},
		\end{split}
	\end{equation}
	where $p_i=F(G(x_i^t))$ is the $K$-dimensional prediction.
	The coefficient $\alpha$ before the pseudo-labeled term is properly designed to grow from 0 to 1 gradually, which can mitigate the noises in the pseudo labels at early iterations to some degree, avoiding the error accumulation.
	
	Since the pseudo labels are not equally confident, in this work, we readily take the maximum predicted probabilities as weights and incorporate them into the standard cross-entropy loss, forming the following objective to adapt the model with unlabeled data (\emph{termed as `pseudo-labeling'}),
	\begin{equation}
		\begin{split}
			\hat{y}_i &= \arg\max\limits_{k} p_{i,k}, \ i=1, 2, \cdots, N_t, \\
			\mathcal{L}_{pl}^{ours} &= - \frac{\lambda}{N_{tu}} \sum\nolimits_{i=1}^{N_{tu}} p_{i,\hat{y}_i} \log p_{i,\hat{y}_i}.
		\end{split}
	\end{equation}
	Different from \cite{lee2013pseudo} where $\alpha$ involves two hyper-parameters, we adopt a simplified linear scheduler for $\lambda$ in this work.
	
	As stated in \cite{lee2013pseudo}, $\mathcal{L}_{pl}$ favors a low-density separation between classes and is in effect equivalent to entropy regularization \cite{grandvalet2005semi} that is employed to reduce the class overlap.
	However, both regularization approaches \cite{lee2013pseudo,grandvalet2005semi} and another recent regularization method \cite{chen2019domain} ignore the structure of unlabeled data and only focus on the instance-wise prediction itself. 
	Considering the data structure especially the diversity among predictions of unlabeled data, Jin \etal \cite{jin2020less} propose to minimize the pair-wise class confusion within a mini-batch of training data.
	Cui \etal \cite{cui2020towards} pursue a lower output matrix rank within a mini-batch to ensure both discriminability and diversity.
	Both approaches have been proven to achieve much better results than vanilla entropy minimization, implying that the structure of the classification output matrix is essential for unlabeled data.
	
	\subsection{Auxiliary Target Domain-Oriented Classifier} 
	In this paper, we propose a new regularization approach called Auxiliary Target Domain-Oriented Classifier (ATDOC) that fully exploits the structure of unlabeled data to get reliable pseudo labels with the presence of domain shift. 
	In particular, ATDOC aims to learn an extra specific classifier $F_t$ for the target domain. 
	However, it is quite challenging to learn $F_t$ without labeled target data. 
	Fortunately, according to prior studies \cite{long2018conditional,pan2020unsupervised}, there exist some source-like samples whose output predictions are reliable, which can be used to help build the classifier proposed here and teach the remaining samples sequentially.
	To avoid the trivial sample selection and alternate training, ATDOC employs a memory module that collects or stores information of all the target samples to generate more accurate pseudo labels.
	In the following, we develop two types of non-parametric target-oriented classifiers and describe the details of them.
	
	\subsubsection{Nearest centroid Classifier (NC)}
	Nearest centroid (NC) classifier is one of the most simple yet powerful classifiers that merely requires the class centroids. Motivated by prior works \cite{liang2019distant,liang2020we} which utilize class centroids to bridge the domain gap for DA problems, we describe the data structure of the target domain with its class centroids.
	Specifically, we introduce a memory bank to store the information of target-oriented class centroids and dynamically generate pseudo-labels with the NC classifier for unlabeled data in each mini-batch.
	
	\textbf{Memory bank update.}
	Since the target data is unlabeled during training, we first obtain the one-hot pseudo labels via $\hat{y}_i = \arg\max\nolimits_{k} p_{i,k}$ and then utilize them to update the centroids in the memory bank via the exponential moving averaging (EMA) strategy, 
	\begin{equation}
		\begin{split}
			%\hat{y}_i &= \arg\max\limits_{k} p_{i,k}, \ i=1, 2, \cdots, N_t, \\
			c_j &= \sum\nolimits_{i \in B_t} \mathbbm{1}_{[j=\hat{y}_i]} G(x_i^t) / \sum\nolimits_{i \in B_t} \mathbbm{1}_{[j=\hat{y}_i]},\\
			c^m_j &= \gamma c_j + (1-\gamma) c^m_j, \ m=1, 2, \cdots, K,
			% \ \forall j \in \{j|\sum\nolimits_{i \in B_t} \mathbbm{1}_{[j=\hat{y}_i]} > 0\},
		\end{split}
	\end{equation}
	where $B_t$ denotes the index set of a min-batch from the target domain, and $\gamma$ is the smoothing parameter fixed to 0.1 as default.
	
	\textbf{Pseudo-labeling.}
	With the class centroids, we readily construct the NC classifier and generate the pseudo labels for each unlabeled datum $x_i^t$ below,
	\begin{equation}
		\hat{y}_i = \arg\min\nolimits_{j=1}^{K} d\left(G(x_i^t),c_j^m\right), \ i=1, 2, \cdots, N_t,
	\end{equation}
	where $d(\cdot,\cdot)$ measures the distance between features and centroids, here we adopt the cosine distance as default.
	Finally, a standard cross-entropy loss is developed as 
	\begin{equation}
		\mathcal{L}_{nc} = - \frac{\lambda}{N_{tu}} \sum\nolimits_{i=1}^{N_{tu}} \log p_{i,\hat{y}_i}.
	\end{equation}

	\subsubsection{Neighborhood Aggregation (NA)}
	Class centroids can only coarsely characterize the domain structure, which fails to depict the local data structure. Therefore, we follow the idea of message passing via neighbors and further develop a neighborhood aggregation (NA) strategy as the classifier.
	Different from the NC classifier, we need to construct a large memory bank to store information e.g. features and soft predictions of each target data.
	
	\textbf{Memory bank update.}
	To avoid ambiguity in the target predictions, we directly sharpen the output predictions $p_i, x_i \in \mathcal{D}_t$ via prediction sharpening and class balancing,
	\begin{equation}
		%\check{p}_{i,k}^{m} = p_{i,k}^{1/T} / \sum\nolimits_i p_{i,k}^{1/T}.
		\check{p}_{i,k}^{m} = p_{i,k}^{1/T} / \sum\nolimits_i p_{i,k}^{1/T}.
		\label{eq:correct}
	\end{equation}
	Note that, the sharpening operation ($T=1/2$) would increase the confidence of each pseudo prediction. Besides, normalizing the sharpened prediction with the overall class-wise vector results in the balancing among different classes over the unlabeled target domain. 
	We do not adopt any moving average strategies to update the values in the memory.
	
	\textbf{Neighborhood aggregation.} 
	The memory module keeps updating every mini-batch, and training a parametric classifier involving extra parameters would be time-consuming.
	To address this, we present a non-parametric neighborhood aggregation strategy as $\mathbf{F}_t$.
	We first retrieve $m$ nearest neighbors from the memory module for each sample in the current mini-batch based on the cosine similarity between their features $G(x_i^t)$ and all the features in the memory bank $f_j^m$. 
	Then, we aggregate corresponding soft predictions of these nearest neighbors by taking the average,
	\begin{equation}
		\hat{q}_i = \frac{1}{m} \sum\nolimits_{j\not= i, j \in \mathcal{N}_i} \check{p}_{j},
	\end{equation}
	where $\mathcal{N}_i$ denotes the index set of neighbors in the memory module for the data point $x_i^t$.
	In this manner, we obtain a new probability prediction via learning on the entire target data.
	Note that our strategy indeed considers the global structure beyond regularization within a mini-batch like \cite{cui2020towards,jin2020less}.
	
	\textbf{Pseudo-labeling.}
	For each unlabeled datum $x_i^t$, we get the pseudo label $\hat{y}_i$ by choosing the category index with the maximum probability prediction $\hat{q}_i$, i.e., $\hat{y}_i = \arg\max_k \hat{q}_{i,k}$.
	Considering different neighborhoods $\mathcal{N}_i$ lie in regions of different densities, it is desirable to assign a larger weight for the target data in a neighborhood of higher density. 
	Intuitively, the larger the maximum value $\hat{q}_{i,\hat{y}_i}$ is, the higher density it will be for the region the datum lies in.
	Thus, we directly utilize $\hat{q}_{i,\hat{y}_i}$ as the confidence (weight) for the pseudo label $\hat{q}_i$.
	Finally, a confidence-weighted cross-entropy loss is imposed on the unlabeled target data as below,
	\begin{equation}
		\mathcal{L}_{na} = - \frac{\lambda}{N_{tu}} \sum\nolimits_{i=1}^{N_{tu}} \hat{q}_{i,\hat{y}_i}  \log p_{i,\hat{y}_i}.
	\end{equation}
	Concerning the labeled data in $\mathcal{D}_s\cup\mathcal{D}_{tl}$, we employ the stand cross-entropy loss with label-smoothing regularization \cite{szegedy2016rethinking}, denoted as $\mathcal{L}_{lsr}^s$ and $\mathcal{L}_{lsr}^t$, respectively.
	Integrating these losses together, we obtain the final objective as follows,
	\begin{equation}
		\mathcal{L} = \mathcal{L}_{lsr}^s(\mathcal{D}_s) + \mathcal{L}_{lsr}^t(\mathcal{D}_{tl}) + \mathcal{L}_{nc/ na}(\mathcal{D}_{tu}).
	\end{equation} 
	Actually, we can readily incorporate $\mathcal{L}_{nc}$ or $\mathcal{L}_{na}$ into other domain alignment methods like CDAN~\cite{long2018conditional} as an additional loss.
	Besides, for SSL methods like MixMatch~\cite{berthelot2019mixmatch}, we just replace $p_{model}(y|u)$ with the \textbf{one-hot encoding} of $\hat{y}$ in the pseudo-labeling step of NC and NA, respectively.

	\section{Experiments}
	\subsection{Setup}
	\textbf{Datasets.}
	Office-31~\cite{saenko2010adapting} is the most widely-used benchmark in the DA field, which consists of 3 different domains in 31 categories: Amazon (\textbf{A}) with 2,817 images, Webcam (\textbf{W}) with 795 images, and DSLR (\textbf{D}) with 498 images. There are 6 transfer tasks for evaluation in total.
	
	Office-Home~\cite{venkateswara2017deep} is another popular benchmark that consists of images from 4 different domains: Artistic (\textbf{Ar}) images, Clip Art (\textbf{Cl}), Product (\textbf{Pr}) images, and Real-World (\textbf{Re}) images, totally around 15,500 images from 65 different categories. All 12 transfer tasks are selected for evaluation.
	
	\setlength{\tabcolsep}{2.0pt}
	\begin{table}[htb]
		\small
		\centering
		\caption{Accuracy (\%) on Office for UDA (ResNet-50). Best (\textbf{\color{red}bold red}), second best (\textit{\color{blue}italic blue}). [$^\dagger$: mean values except D$\leftrightarrow$W]}
		\vspace{3pt}
		\resizebox{0.48\textwidth}{!}{$
			\begin{tabular}{lccccccaa}
			\toprule
			Method  & A$\to$D & A$\to$W & D$\to$A & D$\to$W & W$\to$A & W$\to$D & Avg. & Avg.$^\dagger$\\
			\midrule
			ResNet-50~\cite{he2016deep} & 78.3 & 70.4 & 57.3 & 93.4 & 61.5 & 98.1 & 76.5 & 66.9 \\
			MinEnt~\cite{grandvalet2005semi} & 90.7 & 89.4 & 67.1 & 97.5 & 65.0 & \textbf{\color{red}100.} & 85.0 & 78.1 \\
			MCC~\cite{jin2020less} & 92.1 & 94.0 & 74.9 & 98.5 & 75.3 & \textbf{\color{red}100.} & 89.1 & 84.1 \\
			BNM~\cite{cui2020towards} & 92.2 & 94.0 & 74.9 & 98.5 & 75.3 & \textbf{\color{red}100.} & 89.2 & 84.1 \\
			Pseudo-labeling & 88.7 & 89.1 & 65.8 & 98.1 & 66.6 & 99.6 & 84.7 & 77.6 \\
			ATDOC-NC & 95.2 & 91.6 & 74.6 & \textit{\color{blue}99.1} & 74.7 & \textbf{\color{red}100.} & 89.2 & 84.0 \\  
			ATDOC-NA & 94.4 & 94.3 & 75.6 & 98.9 & 75.2 & 99.6 & 89.7 & 84.9 \\
			\midrule
			CDAN+E~\cite{long2018conditional} & 94.5 & 94.2 & 72.8 & 98.6 & 72.2 & \textbf{\color{red}100.} & 88.7 & 83.4 \\
			+ BSP~\cite{chen2019transferability} & 94.5 & 95.0 & 73.9 & 98.3 & 75.7 & \textbf{\color{red}100.} & 89.6 & 84.8 \\
			+ MCC~\cite{jin2020less} & 94.1 & 94.7 & 75.4 & 99.0 & 75.7 & \textbf{\color{red}100.} & 89.8 & 85.0 \\
			+ BNM~\cite{cui2020towards} & 94.9 & 94.3 & \textit{\color{blue}75.8} & 99.0 & 75.9 & \textbf{\color{red}100.} & \textit{\color{blue}90.0} & \textit{\color{blue}85.2} \\
			+ Pseudo-labeling & 91.5 & 93.1 & 72.5 & 97.8 & 72.7 & 99.8 & 87.9 & 82.4 \\ 
			+ ATDOC-NC & \textit{\color{blue}96.3} & 93.6 & 74.3 & \textit{\color{blue}99.1} & 75.4 & \textbf{\color{red}100.} & 89.8 & 84.9 \\
			+ ATDOC-NA & 95.4 & 94.6 & \textbf{\color{red}77.5} & 98.1 & \textbf{\color{red}77.0} & 99.7 & \textbf{\color{red}90.4} & \textbf{\color{red}86.1} \\		
			\midrule
			MixMatch~\cite{berthelot2019mixmatch} & 88.5 & 84.6 & 63.3 & 96.1 & 65.0 & 99.6 & 82.9 & 75.4 \\
			w/ Pseudo-labeling & 89.0 & 86.0 & 65.8 & 96.2 & 65.6 & 99.6 & 83.7 & 76.6 \\
			w/ ATDOC-NC & 91.3 & 86.4 & 66.0 & 97.4 & 64.4 & 99.4 & 84.1 & 77.0 \\ % momentum = 0.1
			w/ ATDOC-NA & 92.1 & 91.0 & 70.9 & 98.6 & \textit{\color{blue}76.2} & 99.6 & 88.1 & 82.6 \\
			\midrule \midrule
			SAFN+ENT~\cite{xu2019larger} & 90.7 & 90.1 & 73.0 & 98.6 & 70.2 & 99.8 & 87.1 & 81.0 \\
			CRST~\cite{zou2019confidence} & 88.7 & 89.4 & 72.6 & 98.9 & 70.9 & \textbf{\color{red}100.} & 86.8 & 80.4 \\
			SHOT~\cite{liang2020we} & 94.0 & 90.1 & 74.7 & 98.4 & 74.3 & \textit{\color{blue}99.9} & 88.6 & 83.3 \\
			CADA-P~\cite{kurmi2019attending} & 95.6 & \textbf{\color{red}97.0} & 71.5 & \textbf{\color{red}99.3} & 73.1 & \textbf{\color{red}100.} & 89.5 & 84.3 \\
			ATM~\cite{li2020maximum} & \textbf{\color{red}96.4} & \textit{\color{blue}95.7} & 74.1 & \textbf{\color{red}99.3} & 73.5 & \textbf{\color{red}100.} & 89.8 & 84.9 \\
			\bottomrule
			\end{tabular}
			$}
		\label{table:office}
	\end{table}
	
	VisDA-C~\cite{peng2017visda} is a large-scale benchmark used for the \emph{VisDA 2017 Challenge} that consists of 2 very distinct kinds of images from twelve common object classes, i.e., 152,397 synthetic images and 55,388 real images. We focus on the challenging synthetic-to-real transfer task.
	
	DomainNet-126 is a subset of DomainNet~\cite{peng2019moment}, by far the largest UDA dataset with 6 distinct domains and approximately 0.6 million images distributed among 345 categories. Following \cite{saito2019semi}, we pick 126 classes in 4 domains i.e. Real (\textbf{R}), Clipart (\textbf{C}), Painting (\textbf{P}), and Sketch (\textbf{S}) for evaluation. 

	\setlength{\tabcolsep}{3.0pt}
	\begin{table*}[!ht]
		\small
		\centering
		\caption{Per-class accuracy (\%) on VisDA-C validation set using a ResNet-101 backbone.}
		\vspace{3pt}
		\resizebox{0.8\textwidth}{!}{$
		\begin{tabular}{lcccccrcccccra}
			\toprule
			Method  & aero & bike & bus  & car  & horse & knife & mbike & person & plant & skbrd & train & truck & Mean \\
			\midrule
			ResNet-101~\cite{he2016deep} & 67.7 & 27.4 & 50.0 & 61.7 & 69.5 & 13.7 & 85.9 & 11.5 & 64.4 & 34.4 & 84.2 & 19.2 & 49.1 \\
			MinEnt~\cite{grandvalet2005semi} & 88.6 & 29.5 & 82.5 & 75.8 & 88.7 & 16.0 & \textbf{\color{red}93.2} & 63.4 & 94.2 & 40.1 & 87.3 & 12.1 & 64.3 \\
			BNM~\cite{cui2020towards}  & 91.1 & 69.0 & 76.7 & 64.3 & 89.8 & 61.2 & 90.8 & 74.8 & 90.9 & 66.6 & 88.1 & 46.1 & 75.8 \\
			MCC~\cite{jin2020less} & 92.2 & 82.9 & 76.8 & 66.6 & 90.9 & 78.5 & 87.9 & 73.8 & 90.1 & 76.1 & 87.1 & 41.0 & 78.7 \\
			Pseudo-labeling & 90.9 & 74.6 & 73.2 & 55.8 & 89.6 & 64.6 & 86.8 & 68.7 & 90.7 & 64.8 & 89.5 & 47.7 & 74.7 \\
			ATDOC-NC & 91.1 & 60.1 & 78.4 & 72.2 & 88.1 & \textit{\color{blue}97.6} & 86.9 & 55.9 & 79.2 & 64.9 & 88.4 & 31.9 & 74.6 \\ % momentum = 0.1
			ATDOC-NA & 93.7 & 83.0 & 76.9 & 58.7 & 89.7 & 95.1 & 84.4 & 71.4 & 89.4 & 80.0 & 86.7 & 55.1 & 80.3 \\
			\midrule
			CDAN+E~\cite{long2018conditional} & 94.3 & 60.8 & 79.9 & 72.7 & 89.5 & 86.8 & 92.4 & \textit{\color{blue}81.4} & 88.9 & 72.9 & 87.6 & 32.8 & 78.3 \\
			+ BSP~\cite{chen2019transferability} & 93.7 & 58.1 & 80.5 & 69.3 & 89.7 & 86.4 & 92.9 & 78.4 & 88.5 & 74.7 & 88.4 & 33.0 & 77.8 \\
			+ BNM~\cite{cui2020towards} & 93.6 & 68.3 & 78.9 & 70.3 & 91.1 & 82.8 & \textit{\color{blue}93.0} & 78.7 & 90.9 & 76.5 & 89.1 & 40.9 & 79.5 \\
			+ MCC~\cite{jin2020less} & 93.5 & 72.5 & 72.5 & 72.9 & 91.6 & 88.7 & 92.1 & 75.1 & 92.7 & 79.4 & 87.8 & 53.3 & 81.0 \\
			+ Pseudo-labeling & 94.1 & 70.4 & 78.1 & 68.7 & 90.3 & 77.9 & 92.1 & 78.5 & 90.6 & 76.9 & 88.5 & 44.5 & 79.2 \\
			+ ATDOC-NC & 94.3 & 64.0 & 80.1 & 66.2 & 89.1 & 92.1 & 91.0 & 75.4 & 86.9 & 76.3 & 87.2 & 43.4 & 78.8 \\ 
			+ ATDOC-NA & 93.0 & 77.4 & 83.4 & 62.3 & 91.5 & 88.4 & 91.8 & 77.1 & 90.9 & 86.4 & 85.8 & 48.2 & 81.4 \\
			\midrule
			MixMatch~\cite{berthelot2019mixmatch} & 93.9 & 71.8 & \textbf{\color{red}93.5} & \textit{\color{blue}82.1} & 95.3 & 0.7 & 90.8 & 38.1 & 94.5 & \textit{\color{blue}96.0} & 86.3 & 2.2 & 70.4 \\
			w/ Pseudo-labeling & \textit{\color{blue}95.0} & 75.5 & \textit{\color{blue}92.5} & 79.5 & \textbf{\color{red}96.0} & 0.4 & 91.1 & 23.7 & \textbf{\color{red}95.2} & 95.1 & 82.5 & 0.6 & 68.9 \\
			w/ ATDOC-NC & 93.7 & 77.2 & 71.6 & 71.7 & 92.1 & 0.1 & 86.3 & 52.9 & 86.7 & \textbf{\color{red}96.8} & \textbf{\color{red}92.9} & 45.8 & 72.3 \\ 
			w/ ATDOC-NA & \textbf{\color{red}95.3} & \textit{\color{blue}84.7} & 82.4 & 75.6 & \textit{\color{blue}95.8} & \textbf{\color{red}97.7} & 88.7 & 76.6 & 94.0 & 91.7 & \textit{\color{blue}91.5} & \textit{\color{blue}61.9} & \textbf{\color{red}86.3} \\
			\midrule \midrule
			SAFN~\cite{xu2019larger} & 93.6 & 61.3 & 84.1 & 70.6 & 94.1 & 79.0 & 91.8 & 79.6 & 89.9 & 55.6 & 89.0 & 24.4 & 76.1 \\
			CRST~\cite{zou2019confidence} & 88.0 & 79.2 & 61.0 & 60.0 & 87.5 & 81.4 & 86.3 & 78.8 & 85.6 & 86.6 & 73.9 & \textbf{\color{red}68.8} & 78.1 \\
			DTA~\cite{lee2019drop} & 93.7 & 82.2 & 85.6 & \textbf{\color{red}83.8} & 93.0 & 81.0 & 90.7 & \textbf{\color{red}82.1} & \textit{\color{blue}95.1} & 78.1 & 86.4 & 32.1 & 81.5 \\
			SHOT~\cite{liang2020we} & 94.3 & \textbf{\color{red}88.5} & 80.1 & 57.3 & 93.1 & 94.9 & 80.7 & 80.3 & 91.5 & 89.1 & 86.3 & 58.2 & \textit{\color{blue}82.9} \\
			\bottomrule
		\end{tabular}
		$}
	\label{table:visda-c}
	\end{table*}

	\textbf{Implementation\footnote{\url{https://github.com/tim-learn/ATDOC/}} Details.}
	We report the average accuracy over 3 random trials.
	All the methods including domain alignment methods~\cite{long2018conditional,chen2019transferability}, semi-supervised methods~\cite{berthelot2019mixmatch}, and regularization approaches \cite{lee2013pseudo,grandvalet2005semi,chen2019domain,jin2020less,cui2020towards} are implemented based on \textbf{PyTorch} and reverse validation \cite{liang2019exploring,zhong2010cross} is conducted to select hyper-parameters.
	Note that MixMatch~\cite{berthelot2019mixmatch} could be considered as a strong domain adaptation baseline~\cite{rukhovich2019mixmatch}.
	Besides, we select other state-of-the-art UDA approaches~\cite{xu2019larger,zou2019confidence,kurmi2019attending,li2020maximum,lee2019drop,li2020domain} and SSDA approaches~\cite{saito2019semi,jiang2020bidirectional,li2020online,kim2020attract} for further comparison.
	We adopt a linear rampup scheduler from 0 to $\lambda$ for all methods, and $\lambda=0.1$ is fixed for ATDOC-NC, and $\lambda=0.2, m=5$ is fixed for ATDOC-NA throughout this paper.
	We adopt mini-batch SGD to learn the feature encoder by fine-tuning from the ImageNet pre-trained model with the learning rate 0.001, and new layers (bottleneck layer and classification layer) from scratch with the learning rate 0.01.
	We use the suggested training settings in \cite{long2018conditional}, including learning rate scheduler, momentum (0.9), weight decay (1e$^{-3}$), bottleneck size (256), and batch size (36).

	\setlength{\tabcolsep}{1.0pt}
	\begin{table*}[!ht]
		\small
		\centering
		\caption{Accuracy (\%) on Office-Home for closed-set UDA (ResNet-50).}
		\vspace{3pt}
		\resizebox{0.8\textwidth}{!}{$
			\begin{tabular}{lcccccccccccca}
			\toprule
			Method  & Ar$\to$Cl & Ar$\to$Pr & Ar$\to$Re & Cl$\to$Ar & Cl$\to$Pr & Cl$\to$Re & Pr$\to$Ar & Pr$\to$Cl & Pr$\to$Re & Re$\to$Ar & Re$\to$Cl & Re$\to$Pr & Avg.\\
			\midrule
			ResNet-50~\cite{he2016deep} & 44.9 & 66.3 & 74.3 & 51.8 & 61.9 & 63.6 & 52.4 & 39.1 & 71.2 & 63.8 & 45.9 & 77.2 & 59.4 \\
			MinEnt~\cite{grandvalet2005semi} & 51.0 & 71.9 & 77.1 & 61.2 & 69.1 & 70.1 & 59.3 & 48.7 & 77.0 & 70.4 & 53.0 & 81.0 & 65.8 \\
			BNM~\cite{cui2020towards} & 56.7 & 77.5 & 81.0 & 67.3 & 76.3 & 77.1 & 65.3 & 55.1 & 82.0 & 73.6 & 57.0 & 84.3 & 71.1 \\
			MCC~\cite{jin2020less} & 56.3 & 77.3 & 80.3 & 67.0 & 77.1 & 77.0 & 66.2 & 55.1 & 81.2 & 73.5 & 57.4 & 84.1 & 71.0 \\
			Pseudo-labeling & 54.1 & 74.1 & 78.4 & 63.3 & 72.8 & 74.0 & 61.7 & 51.0 & 78.9 & 71.9 & 56.6 & 81.9 & 68.2 \\
			ATDOC-NC & 54.4 & 77.6 & 80.8 & 66.5 & 75.6 & 75.8 & 65.9 & 51.9 & 81.1 & 72.7 & 57.0 & 83.5 & 70.2 \\ 
			ATDOC-NA & 58.3 & \textbf{\color{red}78.8} & \textbf{\color{red}82.3} & \textbf{\color{red}69.4} & \textit{\color{blue}78.2} & \textbf{\color{red}78.2} & 67.1 & 56.0 & \textit{\color{blue}82.7} & 72.0 & 58.2 & \textit{\color{blue}85.5} & \textit{\color{blue}72.2} \\
			\midrule
			CDAN+E~\cite{long2018conditional} & 54.6 & 74.1 & 78.1 & 63.0 & 72.2 & 74.1 & 61.6 & 52.3 & 79.1 & 72.3 & 57.3 & 82.8 & 68.5 \\
			+ BSP~\cite{chen2019transferability}  & 57.1 & 73.4 & 77.5 & 64.2 & 71.8 & 74.3 & 64.0 & 56.7 & 81.0 & 73.4 & 59.1 & 83.3 & 69.6 \\
			+ BNM~\cite{cui2020towards} & 58.1 & 77.2 & 81.1 & 67.5 & 75.3 & 77.2 & 65.5 & \textit{\color{blue}56.8} & 82.6 & 74.1 & \textit{\color{blue}59.9} & 84.6 & 71.7 \\
			+ MCC~\cite{jin2020less} & \textit{\color{blue}58.9} & 77.6 & 80.7 & 67.0 & 75.1 & 77.1 & 65.8 & \textit{\color{blue}56.8} & 82.2 & 73.9 & 59.8 & 84.5 & 71.6 \\
			+ Pseudo-labeling & 57.3 & 76.6 & 79.2 & 66.6 & 74.0 & 76.6 & 66.1 & 53.6 & 81.0 & \textit{\color{blue}74.3} & 58.9 & 84.2 & 70.7 \\
			+ ATDOC-NC & 55.9 & 76.3 & 80.3 & 63.8 & 75.7 & 76.4 & 63.9 & 53.7 & 81.7 & 71.6 & 57.7 & 83.3 & 70.0 \\
			+ ATDOC-NA & \textbf{\color{red}60.2} & 77.8 & \textit{\color{blue}82.2} & \textit{\color{blue}68.5} & \textbf{\color{red}78.6} & 77.9 & \textbf{\color{red}68.4} & \textbf{\color{red}58.4} & \textbf{\color{red}83.1} & \textbf{\color{red}74.8} & \textbf{\color{red}61.5} & \textbf{\color{red}87.2} & \textbf{\color{red}73.2} \\
			\midrule\midrule
			SAFN~\cite{xu2019larger} & 52.0 & 71.7 & 76.3 & 64.2 & 69.9 & 71.9 & 63.7 & 51.4 & 77.1 & 70.9 & 57.1 & 81.5 & 67.3 \\
			CADA-P~\cite{kurmi2019attending} & 56.9 & 76.4 & 80.7 & 61.3 & 75.2 & 75.2 & 63.2 & 54.5 & 80.7 & 73.9 & \textbf{\color{red}61.5} & 84.1 & 70.2 \\
			DCAN~\cite{li2020domain} & 54.5 & 75.7 & 81.2 & 67.4 & 74.0 & 76.3 & \textit{\color{blue}67.4} & 52.7 & 80.6 & 74.1 & 59.1 & 83.5 & 70.5 \\
			SHOT~\cite{liang2020we} & 57.1 & \textit{\color{blue}78.1} & 81.5 & 68.0 & \textit{\color{blue}78.2} & \textit{\color{blue}78.1} & \textit{\color{blue}67.4} & 54.9 & 82.2 & 73.3 & 58.8 & 84.3 & 71.8 \\
			\bottomrule
			\end{tabular}
			$}
		\label{table:home}
	\end{table*}  
	
	\setlength{\tabcolsep}{3.0pt}
	\begin{table*}[!ht]
		\small
		\centering
		\caption{Accuracy (\%) on DomainNet-126 for Semi-supervised DA (SSDA) using a ResNet-34 backbone.}
		\vspace{3pt}
		\resizebox{0.9\textwidth}{!}{$
			\begin{tabular}{lccccccccccccccaa}
			\toprule
			Method  & \multicolumn{2}{c}{C $\to$ S} & \multicolumn{2}{c}{P $\to$ C} & \multicolumn{2}{c}{P $\to$ R} & \multicolumn{2}{c}{R $\to$ C} & \multicolumn{2}{c}{R $\to$ P} & \multicolumn{2}{c}{R $\to$ S} & \multicolumn{2}{c}{S $\to$ P} & \multicolumn{2}{c}{Average}\\
			& 1-shot & 3-shot & 1-shot & 3-shot & 1-shot & 3-shot & 1-shot & 3-shot & 1-shot & 3-shot & 1-shot & 3-shot & 1-shot & 3-shot & 1-shot & 3-shot\\
			\midrule
			ResNet-34~\cite{he2016deep} & 54.8 & 57.9 & 59.2 & 63.0 & 73.7 & 75.6 & 61.2 & 63.9 & 64.5 & 66.3 & 52.0 & 56.0 & 60.4 & 62.2 & 60.8 & 63.6 \\
			MinEnt~\cite{grandvalet2005semi} & 56.3 & 61.5 & 67.7 & 71.2 & 76.0 & 78.1 & 66.1 & 71.6 & 68.9 & 70.4 & 60.0 & 63.5 & 62.9 & 66.0 & 65.4 & 68.9 \\ 
			MCC~\cite{jin2020less}  & 56.8 & 60.5 & 62.8 & 66.5 & 75.3 & 76.5 & 65.5 & 67.2 & 66.9 & 68.1 & 57.6 & 59.8 & 63.4 & 65.0 & 64.0 & 66.2 \\
			BNM~\cite{cui2020towards} & 58.4 & 62.6 & 69.4 & 72.7 & 77.0 & 79.5 & 69.8 & 73.7 & 69.8 & 71.2 & 61.4 & 65.1 & 64.1 & 67.6 & 67.1 & 70.3 \\
			Pseudo-labeling & 62.5 & 64.5 & 67.6 & 70.7 & 78.3 & 79.3 & 70.9 & 72.9 & 69.2 & 70.7 & 62.0 & 64.8 & 67.0 & 68.6 & 68.2 & 70.2 \\
			ATDOC-NC & 58.1 & 62.2 & 65.8 & 70.2 & 76.9 & 78.7 & 69.2 & 72.3 & 69.8 & 70.6 & 60.4 & 65.0 & 65.5 & 68.1 & 66.5 & 69.6 \\ % momentum = 0.1
			ATDOC-NA & \textbf{\color{red}65.6} & \textbf{\color{red}66.7} & \textit{\color{blue}72.8} & 74.2 & \textbf{\color{red}81.2} & \textbf{\color{red}81.2} & \textbf{\color{red}74.9} & \textbf{\color{red}76.9} & \textbf{\color{red}71.3} & \textbf{\color{red}72.5} & \textit{\color{blue}65.2} & 64.6 & \textbf{\color{red}68.7} & \textbf{\color{red}70.8} & \textbf{\color{red}71.4} & \textbf{\color{red}72.4} \\
			\midrule
			MixMatch~\cite{berthelot2019mixmatch} & 59.3 & 62.7 & 66.7 & 68.7 & 74.8 & 78.8 & 69.4 & 72.6 & 67.8 & 68.8 & 62.5 & 65.6 & 66.3 & 67.1 & 66.7 & 69.2 \\
			w/ Pseudo-labeling & 59.6 & 62.6 & 67.5 & 69.6 & 74.8 & 78.6 & 70.0 & 73.0 & 68.6 & 69.3 & 63.2 & 65.9 & 66.6 & 67.3 & 67.2 & 69.5 \\
			w/ ATDOC-NC & 60.2 & 63.4 & 65.2 & 69.5 & 75.0 & 78.9 & 68.4 & 73.0 & 68.7 & 70.1 & 60.9 & 64.5 & 65.3 & 67.1 & 66.2 & 69.5 \\
			w/ ATDOC-NA & \textit{\color{blue}64.6} & \textit{\color{blue}65.9} & 70.7 & 72.2 & \textit{\color{blue}80.3} & \textit{\color{blue}80.8} & \textit{\color{blue}74.0} & 75.2 & 70.2 & 71.2 & \textbf{\color{red}65.7} & \textit{\color{blue}67.7} & \textit{\color{blue}68.5} & \textit{\color{blue}69.4} & \textit{\color{blue}70.6} & \textit{\color{blue}71.8} \\
			\midrule
			MME~\cite{saito2019semi} & 56.3 & 61.8 & 69.0 & 71.7 & 76.1 & 78.5 & 70.0 & 72.2 & 67.7 & 69.7 & 61.0 & 61.9 & 64.8 & 66.8 & 66.4 & 68.9 \\
			BiAT~\cite{jiang2020bidirectional} & 57.9 & 61.5 & 71.6 & \textit{\color{blue}74.6} & 77.0 & 78.6 & 73.0 & 74.9 & 68.0 & 68.8 & 58.5 & 62.1 & 63.9 & 67.5 & 67.1 & 69.7 \\
			Meta-MME~\cite{li2020online} & - & 62.8 & - & 72.8 & - & 79.2 & - & 73.5 & - & 70.3 & - & 63.8 & - & 68.0 & - & 70.1 \\
			APE~\cite{kim2020attract} & 56.7 & 63.1 & \textbf{\color{red}72.9} & \textbf{\color{red}76.7} & 76.6 & 79.4 & 70.4 & \textit{\color{blue}76.6} & \textit{\color{blue}70.8} & \textit{\color{blue}72.1} & 63.0 & \textit{\color{red}67.8} & 64.5 & 66.1 & 67.6 & 71.7 \\
			\bottomrule
			\end{tabular}
			$}
		\label{table:ssda:domainnet}
	\end{table*}        
	
	\setlength{\tabcolsep}{2.0pt}
	\begin{table*}[ht]
		\small
		\centering
		\caption{Accuracy (\%) on Office-Home for Partial-set UDA (PDA) using a ResNet-50 backbone.}
		\vspace{3pt}
		\resizebox{0.78\textwidth}{!}{$
			\begin{tabular}{lcccccccccccca}
			\toprule
			Method  & Ar$\to$Cl & Ar$\to$Pr & Ar$\to$Re & Cl$\to$Ar & Cl$\to$Pr & Cl$\to$Re & Pr$\to$Ar & Pr$\to$Cl & Pr$\to$Re & Re$\to$Ar & Re$\to$Cl & Re$\to$Pr & Avg. \\
			\midrule
			ResNet-50~\cite{he2016deep} & 43.5 & 67.8 & 78.9 & 57.5 & 56.2 & 62.2 & 58.1 & 40.7 & 74.9 & 68.1 & 46.1 & 76.3 & 60.9 \\
			MinEnt~\cite{grandvalet2005semi} & 45.7 & 73.3 & 81.6 & 64.6 & 66.2 & 73.0 & 66.0 & 52.4 & 78.7 & 74.8 & 56.7 & 80.8 & 67.8 \\ 
			MCC~\cite{jin2020less} & 54.1 & 75.3 & 79.5 & 63.9 & 66.3 & 71.8 & 63.3 & 55.1 & 78.0 & 70.4 & 55.7 & 76.7 & 67.5 \\
			%MCC~\cite{jin2020less} & 63.1 & 80.8 & 86.0 & 70.8 & 72.1 & 80.1 & 75.0 & 60.8 & 85.9 & 78.6 & 65.2 & 82.8 & 75.1 \\
			BNM~\cite{cui2020towards} & 54.6 & 77.2 & 81.1 & 64.9 & 67.9 & 72.8 & 62.6 & 55.7 & 79.4 & 70.5 & 54.7 & 77.6 & 68.2 \\
			Pseudo-labeling & 51.9 & 70.7 & 77.5 & 61.7 & 62.4 & 67.8 & 62.9 & 54.1 & 73.8 & 70.4 & 56.7 & 75.0 & 65.4 \\
			ATDOC-NC & 59.5 & \textbf{\color{red}80.3} & \textit{\color{blue}83.8} & \textit{\color{blue}71.8} & \textit{\color{blue}71.6} & 79.7 & 70.6 & \textit{\color{blue}59.4} & 82.2 & \textbf{\color{red}78.4} & \textbf{\color{red}61.1} & 81.5 & \textit{\color{blue}73.3} \\ 
			ATDOC-NA & \textit{\color{blue}60.1} & 76.9 & \textbf{\color{red}84.5} & \textbf{\color{red}72.8} & 71.2 & \textbf{\color{red}80.9} & \textbf{\color{red}73.9} & \textbf{\color{red}61.8} & 83.8 & 77.3 & \textit{\color{blue}60.4} & 80.4 & \textbf{\color{red}73.7} \\
			\midrule
			ETN~\cite{cao2019learning} & 59.2 & 77.0 & 79.5 & 62.9 & 65.7 & 75.0 & 68.3 & 55.4 & \textit{\color{blue}84.4} & 75.7 & 57.7 & \textit{\color{blue}84.5} & 70.5 \\
			SAFN~\cite{xu2019larger} & 58.9 & 76.3 & 81.4 & 70.4 & \textit{\color{blue}73.0} & 77.8 & \textit{\color{blue}72.4} & 55.3 & 80.4 & 75.8 & \textit{\color{blue}60.4} & 79.9 & 71.8 \\
			RTNet$_{adv}$ \cite{chen2020selective} & \textbf{\color{red}63.2} & \textit{\color{blue}80.1} & 80.7 & 66.7 & 69.3 & 77.2 & 71.6 & 53.9 & \textbf{\color{red}84.6} & \textit{\color{blue}77.4} & 57.9 & \textbf{\color{red}85.5} & 72.3 \\
			\bottomrule
			\end{tabular}
			$}
		\label{table:home-pda}
		\vspace{-15pt}
	\end{table*}
	
	\subsection{Results}
	\noindent \textbf{$\triangleright$ Results of Closed-set Unsupervised DA (UDA)}.
	We use three datasets as introduced above for vanilla UDA tasks, with results shown in Tables~\ref{table:office}$\sim$\ref{table:home}.
	On the small-sized Office-31 dataset, we first study different SSL regularization approaches when integrated with the source classification loss only.
	It is obvious that both BNM~\cite{cui2020towards} and MCC~\cite{jin2020less} consistently perform better than instance-wise regularization methods like MinEnt~\cite{grandvalet2005semi}, which verifies the importance of local diversity. 
	Besides, ATDOC-NC achieves competitive results against BNM, outperforming MinEnt and Pseudo-labeling.
	ATDOC-NA outperforms BNM in 4 out of 6 tasks, obtaining the best average accuracy.
	When combined with one popular UDA method i.e. CDAN+E~\cite{long2018conditional}, the average accuracy increases accordingly, and ATDOC-NA still performs the best.
	Since Office-31 is relatively small, MixMatch \cite{berthelot2019mixmatch} performs worse than CDAN+E.
	Using pseudo labels provided by ATDOC-NC and ATODC-NA, MixMatch obtains boosted performance. 
	Besides, ATDOC-NA achieves competitive performance with state-of-the-art UDA methods like ATM~\cite{li2020maximum} \textit{without any explicit feature-level alignment}. 
	
	% 	For VisDA-C and Office-Home, we compare the performance between BNM, MCC, and ATDOC-NA with or without domain alignment, respectively.
	As shown in Table~\ref{table:visda-c}, ATDOC-NA clearly performs better than BNM and MCC w.r.t. mean accuracy for both situations.
	Note, ATDOC-NA combined with MixMatch obtains the state-of-the-art mean accuracy 86.3\% for VisDA-C, which outperforms recent UDA methods~\cite{xu2019larger,zou2019confidence,lee2019drop,liang2020we}.
	Taking a closer look at Table~\ref{table:home}, we observe similar results for Office-Home that ATDOC-NA beats ATDOC-NC and BNM in terms of the average accuracy.
	Integrated with the feature-level method\textemdash CDAN+E, ATDOC-NA clearly beats the state-of-the-art approach, SHOT \cite{liang2020we}.
	
	\begin{figure*}[h]
		\centering
		\scriptsize
		\renewcommand\arraystretch{1.0}
		\begin{tabular}{cc}
			\includegraphics[width=0.245\linewidth, trim=110 230 105 180,clip]{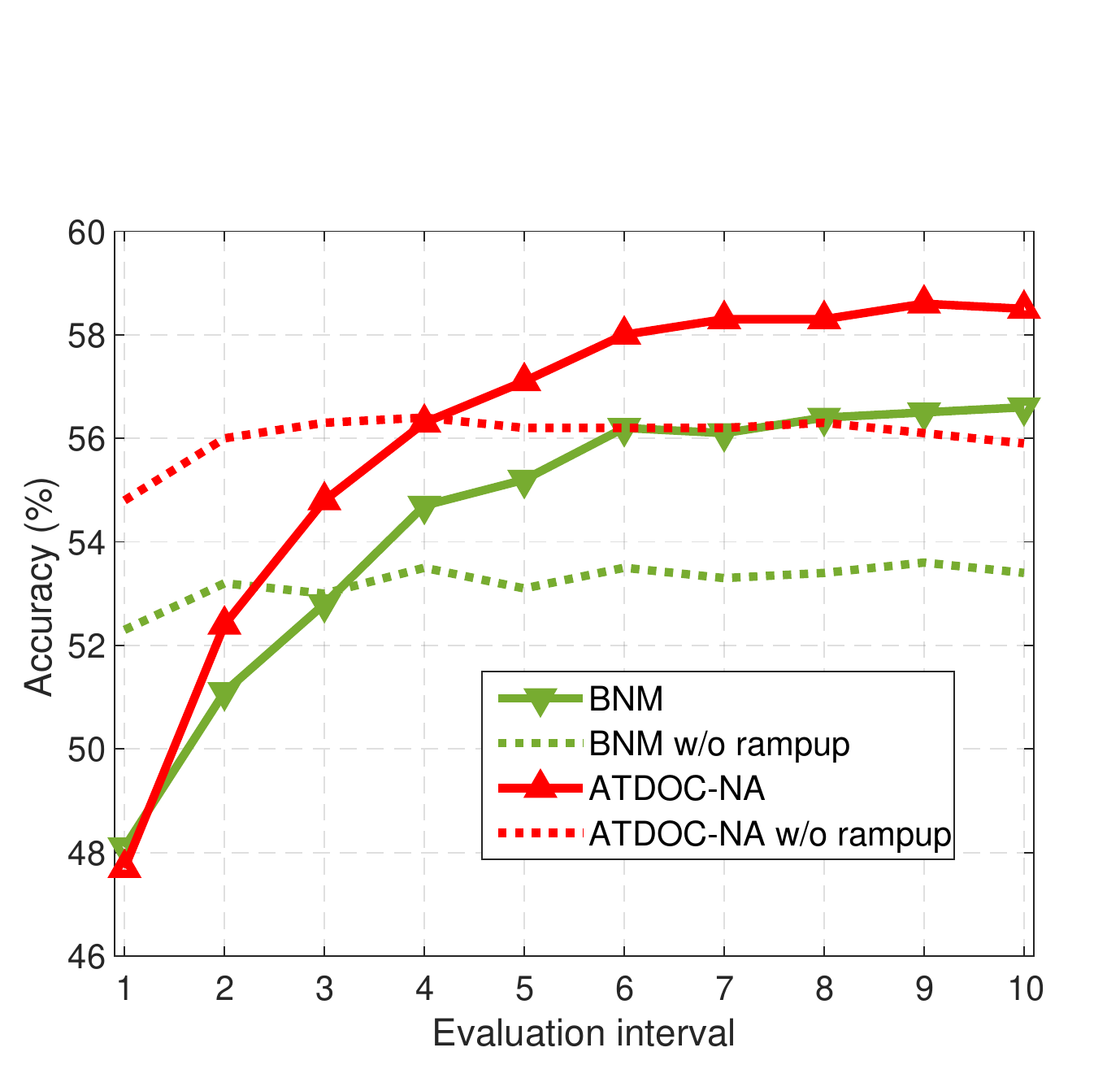} & 
			\includegraphics[width=0.245\linewidth, trim=110 230 105 180,clip]{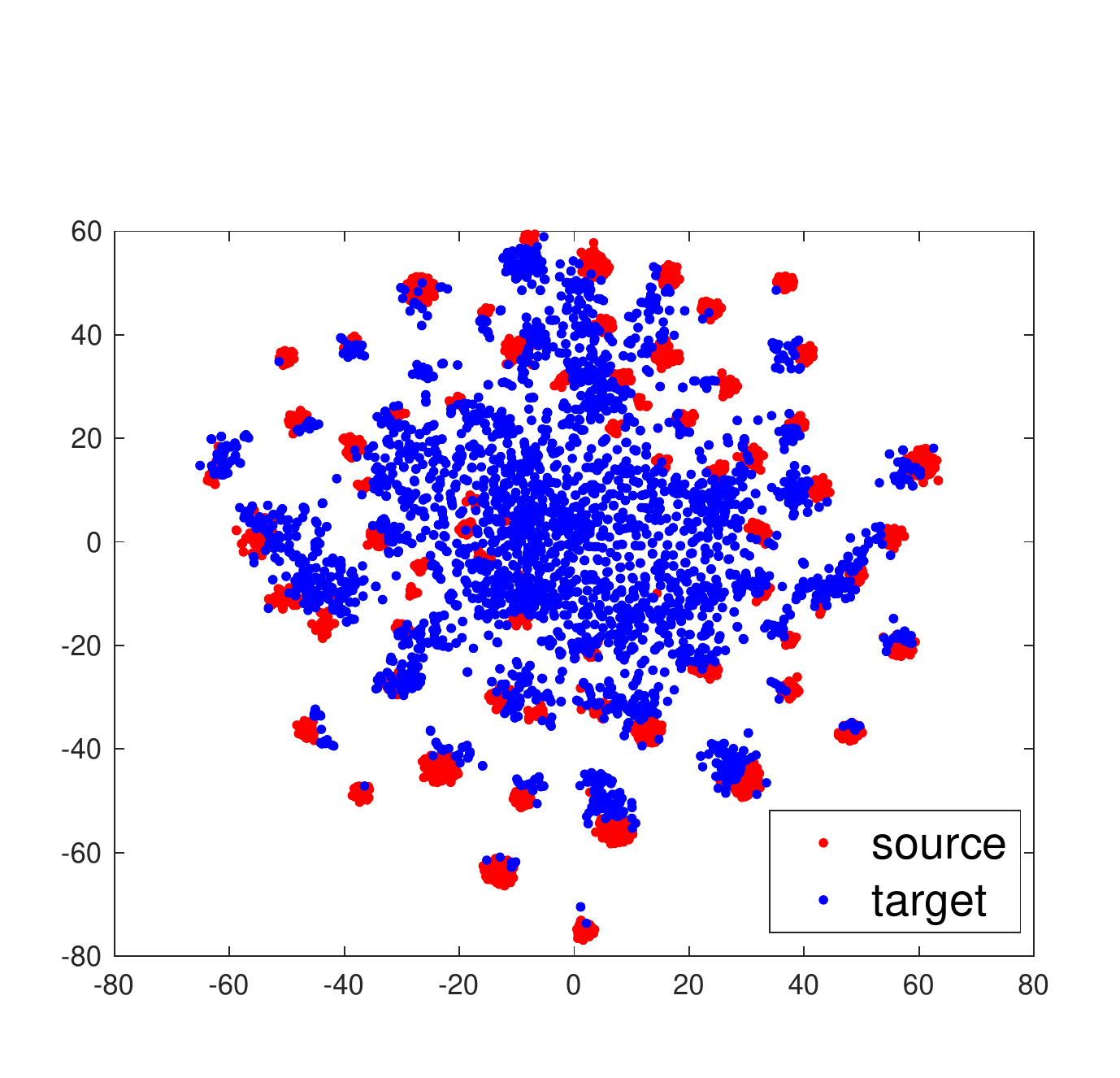} 
			\includegraphics[width=0.245\linewidth, trim=110 230 105 180,clip]{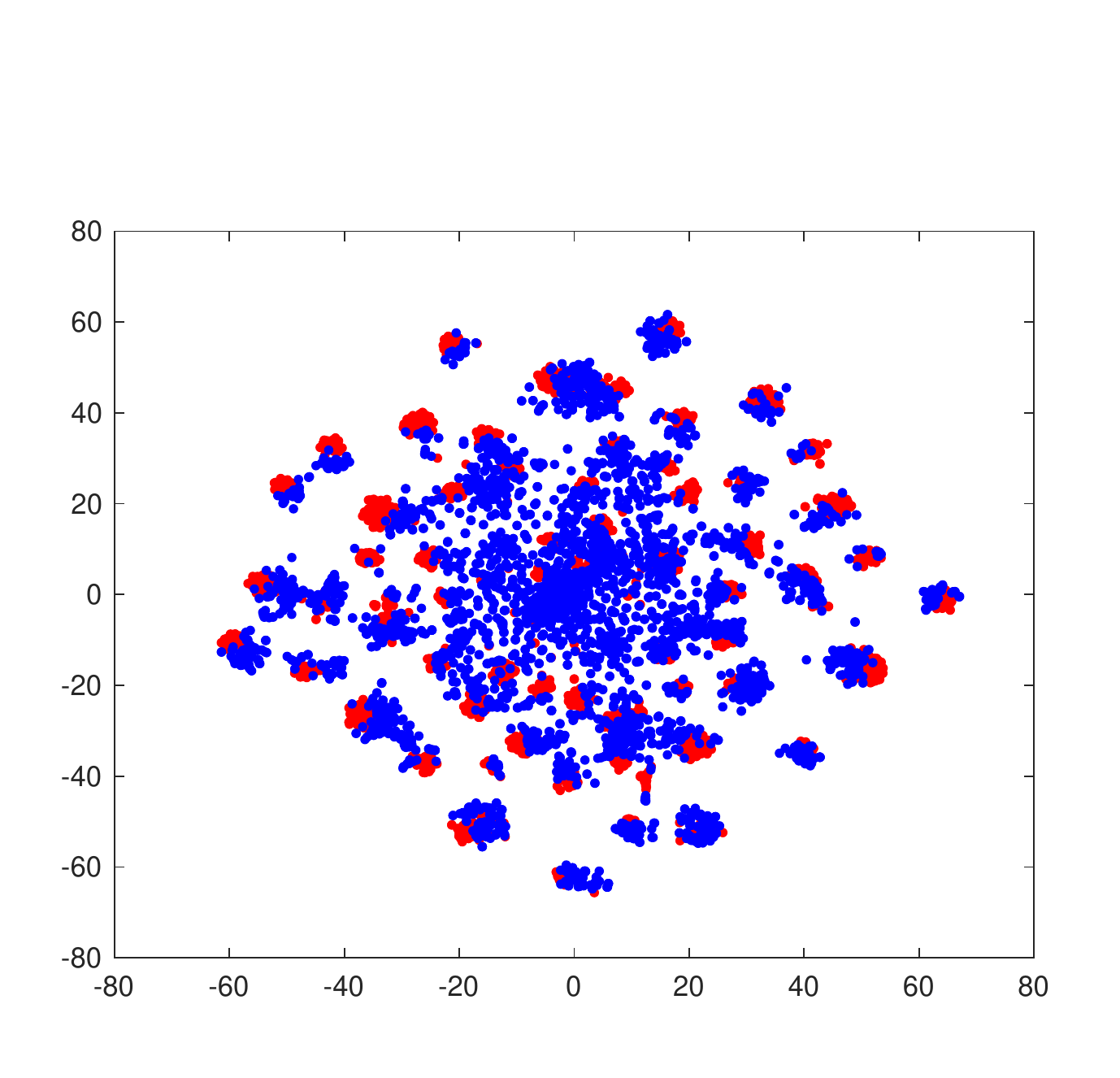} 
			\includegraphics[width=0.245\linewidth, trim=110 230 105 180,clip]{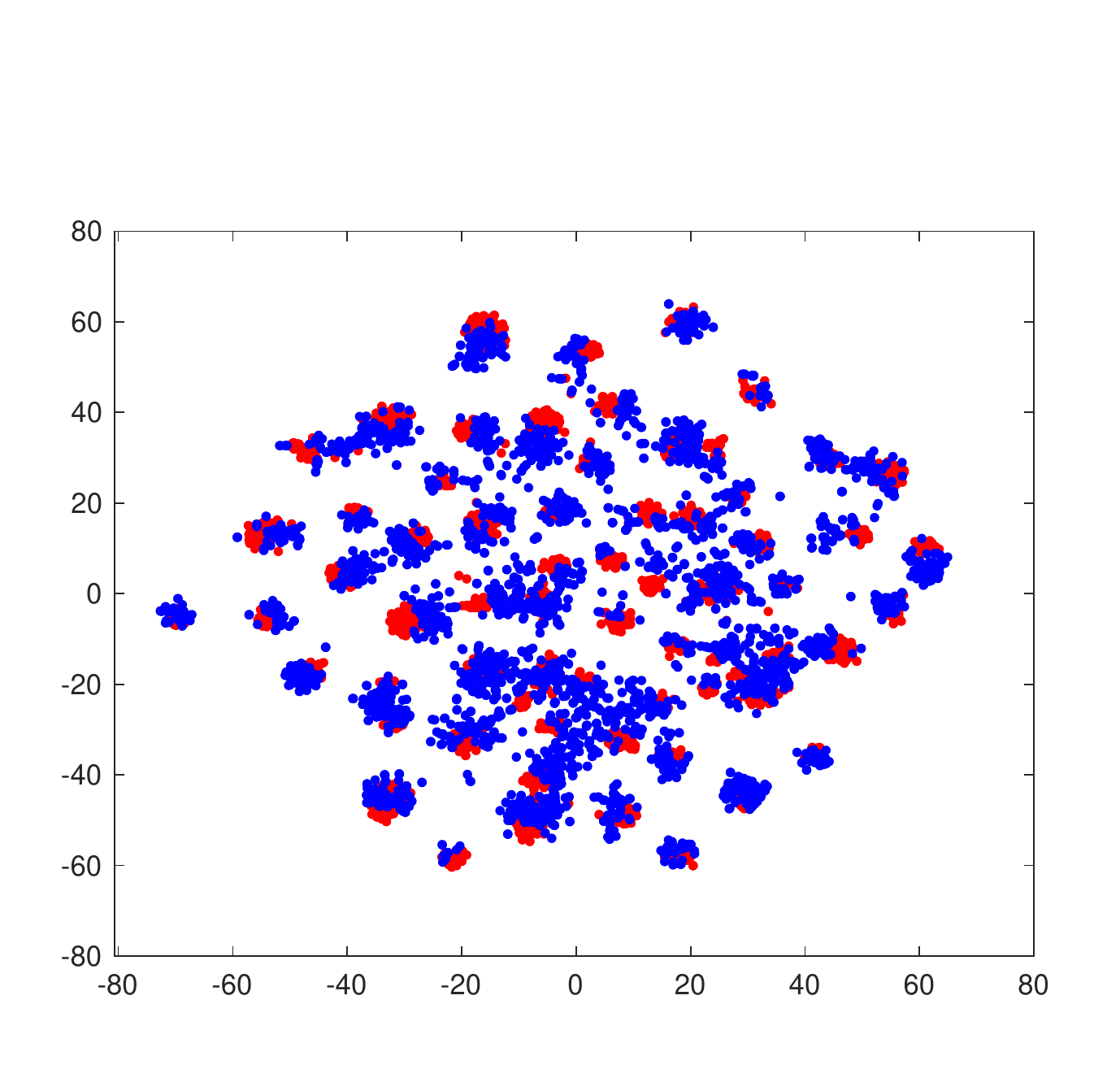} \\
			(a) Convergence & (b) t-SNE visualizations of features learned by Source-only (ResNet-50), Pseudo-labeling, and ATDOC-NA \\
		\end{tabular}
		\vspace{3pt}
		\caption{For Ar$\to$Cl task on Office-Home, (a) shows the convergence and (b) shows the t-SNE visualizations. ({\color{red}red}: Ar, {\color{blue}blue}: Cl)}
		\label{fig:par}
	\end{figure*}
	
	\noindent \textbf{$\triangleright$ Results of Semi-supervised DA (SSDA)}.
	We follow the settings in MME~\cite{saito2019semi} and evaluate SSDA methods on the DomainNet-126 dataset.
	There exist two SSDA protocols in which each class in the target domain has one or three labeled data points, respectively. 
	As shown in Table~\ref{table:ssda:domainnet}, ATDOC-NA outperforms both BNM and MCC for both protocols, and MixMatch also benefits from the incorporation of ATDOC.
	Comparing the results of ATDOC-NA under 1-shot and 3-shot, we find the difference between them is relatively small, implying that ATDOC-NA can fully exploit the unlabeled data to compensate for the scarcity of labeled data.
	Moreover, compared with prior state-of-the-art SSDA results reported in APE~\cite{kim2020attract}, both ATDOC-NA and its combination with MixMatch achieve better performance for both protocols.
	
	\noindent \textbf{$\triangleright$ Results of Partial-set UDA (PDA)}.
	We adopt the standard partial-set UDA setting as \cite{cao2019learning,liang2020balanced} on the Office-Home dataset, in which only data of the first 25 classes exist in the target domain.
	As can be seen from Table~\ref{table:home-pda}, ATDOC-NC behaves competitively against ATDOC-NA, this may be because class centroids are somewhat global-level resisting the noise. 
	Faced with the challenging asymmetric label space in PDA, BNM does not clearly outperform MinEnt anymore, but the adaptation results of ATDOC are still fairly promising.
	Compared with a recent PDA approach, RTNet$_{adv}$ \cite{chen2020selective}, ATDOC-NA obtains a better average accuracy.
	
	\setlength{\tabcolsep}{2.0pt}
	\begin{table}[ht]
		\small
		\centering
		\caption{Accuracy (\%) on Office-Home and DomainNet-126 for scarce-labeled SSL (ResNet-50). }
		\vspace{3pt}
		\resizebox{0.47\textwidth}{!}{$
			\begin{tabular}{lccccaccccca}
			\toprule
			Dataset  & \multicolumn{5}{c}{Office-Home} & & \multicolumn{5}{c}{DomainNet-126} \\
			\cmidrule{2-6}    \cmidrule{8-12}
			Method & Ar & Cl & Pr & Re & Avg.  & & C & P & R & S & Avg. \\
			\midrule
			ResNet-50~\cite{he2016deep} & 48.7 & 42.1 & 68.9 & 66.6 & 56.6 &  & 41.6 & 46.2 & 66.4 & 33.3 & 46.9 \\
			MinEnt~\cite{grandvalet2005semi} & 51.7 & 44.5 & 72.4 & 68.9 & 59.4 &  & 43.8 & 48.6 & 68.8 & 35.4 & 49.2 \\
			MCC~\cite{jin2020less} & 58.9 & \textit{\color{blue}47.7} & \textit{\color{blue}77.4} & 74.3 & 64.6 &  & 45.7 & 49.5 & 70.9 & 38.5 & 51.2 \\
			BNM~\cite{cui2020towards} & \textit{\color{blue}59.0} & 46.0 & 76.5 & 71.5 & 63.2 &  & 44.8 & 47.2 & 69.9 & 35.5 & 49.4 \\
			Pseudo-labeling & 47.3 & 41.4 & 71.4 & 66.1 & 56.6 &  & 41.0 & 46.3 & 72.5 & 33.1 & 48.2 \\
			ATDOC-NC & 56.0 & 43.4 & 76.6 & 72.9 & 62.2 && 45.6 & 51.5 & 72.2 & 35.2 & 51.1 \\ % momentum = 0.1
			ATDOC-NA & \textbf{\color{red}59.1} & 46.6 & \textbf{\color{red}78.4} & \textbf{\color{red}75.9} & \textbf{\color{red}65.0} && \textbf{\color{red}54.7} & \textbf{\color{red}60.0} & \textbf{\color{red}75.5} & \textit{\color{blue}38.6} & \textbf{\color{red}57.2} \\
			\midrule
			MixMatch~\cite{berthelot2019mixmatch} & 52.2 & 41.9 & 73.1 & 69.1 & 59.1 && 41.2 & 38.7 & 64.3 & 34.2 & 44.6 \\
			w/ Pseudo-labeling & 53.4 & 42.5 & 72.6 & 69.5 & 59.5 && 40.4 & 39.1 & 64.7 & 34.1 & 44.6 \\
			w/ ATDOC-NC & 54.6 & 44.4 & 72.7 & 71.0 & 60.7 && 41.2 & 38.3 & 63.4 & 34.4 & 44.3 \\  % momentum = 0.1
			w/ ATDOC-NA & 56.4 & \textbf{\color{red}48.3} & 74.7 & \textit{\color{blue}75.2} & \textit{\color{blue}63.6} && \textit{\color{blue}49.9} & \textit{\color{blue}51.4} & \textit{\color{blue}72.8} & \textbf{\color{red}40.8} & \textit{\color{blue}53.7} \\
			\bottomrule
			\end{tabular}
			$}
		\label{table:ssl}
	\end{table}  
	
	\setlength{\tabcolsep}{2.0pt}
	\begin{table}[!htb]
		\small
		\centering
		\caption{Ablation study.}
		\vspace{3pt}
		\resizebox{0.46\textwidth}{!}{$
			\begin{tabular}{llll}
			\toprule
			Method/ Dataset & & Office-31 & VisDA-C \\
			\midrule
			ATDOC-NA (\emph{\textbf{default}, $T=0.5, m=5,\lambda=0.2$}) & & 89.7 ({\textcolor{black}{--}}) & 80.3 ({\textcolor{black}{--}}) \\
			\midrule 
			ATDOC-NA w/o weight $\check{q}_{i,\check{y}_i}$ & & 89.4 ({\textcolor{green}{$\downarrow$}}) & 79.2 ({\textcolor{green}{$\downarrow$}}) \\
			ATDOC-NA w/ temperature $T=1$ &  & 85.8 ({\textcolor{green}{$\downarrow$}}) & 65.6 ({\textcolor{green}{$\downarrow$}}) \\
			ATDOC-NA (w/ \textbf{source memory}) &  & 89.4 ({\textcolor{green}{$\downarrow$}}) & 80.9 ({\textcolor{red}{$\uparrow$}}) \\
			\midrule
			ATDOC-NA w/ neighborhood size $m=1$ &  & 84.7 ({\textcolor{green}{$\downarrow$}}) & 79.9 ({\textcolor{green}{$\downarrow$}}) \\
			ATDOC-NA w/ neighborhood size $m=3$ &  & 87.9 ({\textcolor{green}{$\downarrow$}}) & 80.0 ({\textcolor{green}{$\downarrow$}})\\
			\midrule
			ATDOC-NA w/ parameter $\lambda=0.1$ &  & 90.2 ({\textcolor{red}{$\uparrow$}}) & 78.8 ({\textcolor{green}{$\downarrow$}})\\
			ATDOC-NA w/ parameter $\lambda=0.3$ &  & 89.2 ({\textcolor{green}{$\downarrow$}}) & 80.9 ({\textcolor{red}{$\uparrow$}}) \\
			\midrule
			\midrule
			ATDOC-NC (\emph{\textbf{default}, $\lambda=0.1$}) &  & 89.2 ({\textcolor{black}{--}}) & 74.6 ({\textcolor{black}{--}}) \\
			ATDOC-NC w/ parameter $\lambda=0.2$ &  & 88.6 ({\textcolor{green}{$\downarrow$}}) & 76.4 ({\textcolor{red}{$\uparrow$}}) \\
			ATDOC-NC w/ parameter $\lambda=0.3$ &  & 88.3 ({\textcolor{green}{$\downarrow$}}) & 77.0 ({\textcolor{red}{$\uparrow$}}) \\
			ATDOC-NC (w/ \textbf{source memory}) &  & 87.4 ({\textcolor{green}{$\downarrow$}}) & 66.6 ({\textcolor{green}{$\downarrow$}}) \\	
			\bottomrule
			\end{tabular}
			$}
		\label{tab:abl}
	\end{table}

	\noindent \textbf{$\triangleright$ Results of Scare-labeled SSL}.
	We further study ATDOC in a special SSL case without the domain shift where labeled samples are very scarce.
	For simplicity, we adopt the same three-shot setting in the aforementioned SSDA, but take labeled target data as the labeled set and unlabeled target data as the unlabeled set, forming the scarce-labeled SSL task. 
	As shown in Table~\ref{table:ssl},  ATDOC-NA performs the best on both Office-Home and DomainNet-126.
	For such a scarce-labeled SSL task, MixMatch performs badly. 
	The reason may be that labeled data are quite scarce, resulting in low-quality pseudo labels and thus bringing much noise in the following MixUp step.
	ATDOC-NA improves the quality of pseudo labels and significantly boosts the performance when replacing the label guessing process in MixMatch. 
	Benefited from a large amount of unlabeled data, ATDOC-NA outperforms MCC for SSL tasks on DomainNet-126 with a larger margin than that on Office-Home.
	
	\subsection{Model Analysis}
	We study the convergence of ATDOC-NA and the ramp-up of $\lambda$, and make comparisons with BNM in Fig.~\ref{fig:par}(a). 
	Comparing both methods with or without the ramp-up, it is easy to verify the effectiveness of linear ramp-up.
	Since the pseudo labels or original classifier outputs in the early stage are not reliable enough, using a ramp-up to progressively increase the regularization weight is desirable for both ATDOC-NA and BNM.
	Besides, with the iteration number increasing, the accuracy of ATDOC-NA grows up and converges at last.
	Furthermore, we employ the t-SNE visualization~\cite{maaten2008visualizing} in Fig.~\ref{fig:par}(b) to show whether features from different domains are well aligned even without explicit domain alignment.
	Compared with source-only (ResNet-50) and pseudo-labeling, features from both domains learned by ATDOC-NA are semantically aligned and more favorable.

	We further conduct ablation on Office-31 and VisDA-C for UDA and show average accuracy in Table~\ref{tab:abl}.
	Comparing results in the first three rows, we find both weighting and class-balancing sharpening strategies are effective.
	Besides, we study the neighborhood size $m$ for ATDOC-NA and find a larger value of $m$ can bring better performance.
	In particular, on the small Office-31 dataset, using $m=1$ is quite risky and achieves worse results.
	Regarding another parameter $\lambda$, we discover $\lambda=0.2$ is a suitable choice for both datasets.
	For the large-scale VisDA-C dataset, the learned pseudo labels are more reliable, so a large value of $\lambda$ is beneficial.
	In addition, we investigate the effects of different parameters $\lambda$ for ATDOC-NC.
	For a large-scale dataset, increasing $\lambda$ is a better choice while decreasing $\lambda$ is suitable for small-scale datasets.
	At last, we study the incorporation of source features in the memory bank and find it always degrades the performance, verifying the effectiveness of the target-oriented classifier.
	
	\section{Conclusion}
	We presented ATDOC, a new regularization approach to address the dataset shift for DA tasks.
	Despite the simplicity, extensive experiments demonstrated that ATDOC-NA outperforms both feature-level domain alignment methods and other regularization methods with consistent margins on UDA, SSDA, PDA, and even scarce-labeled SSL tasks. 
	In the future, we would like to extend ATDOC to other challenging transfer tasks like universal DA~\cite{saito2020universal,you2019universal} and dense labeling tasks like semantic segmentation~\cite{tsai2018learning,chen2019crdoco}.
	
	\section{Correction}
	Now we correct the typo in Eq.~(\ref{eq:correct}) of the camera-ready version. All the previous experimental results were produced by Eq.~(\ref{eq:correct}) in this new version, thus needing no further changes. We would like to specially thank the github user \textbf{\color{red}@lyxok1} for pointing out this typo under our repository.
	
	\section{Acknowledgment}
	This work was partially supported by AISG-100E-2019-035, MOE2017-T2-2-151, NUS\_ECRA\_FY17\_P08 and CRP20-2017-0006.
	
	{\small
		\bibliographystyle{ieee_fullname}
		\balance
		\bibliography{egbib}
	}
	
\end{document}